\documentclass[10pt,twocolumn,letterpaper]{article}

\usepackage[pagenumbers]{cvpr} 

\definecolor{cvprblue}{rgb}{0.21,0.49,0.74}
\usepackage[pagebackref,breaklinks,colorlinks,allcolors=cvprblue]{hyperref}

\usepackage{amsmath, amssymb, amsfonts} 

\def\paperID{11197} 
\def\confName{CVPR}
\def\confYear{2026}

\title{PathMamba: A Hybrid Mamba-Transformer for Topologically Coherent Road Segmentation in Satellite Imagery}

\author{\large
Jules Decaestecker\thanks{decaestecker.jules@gmail.com} \qquad
Nicolas Vigne\thanks{nicolas.vigne@thalesgroup.com} \\
\\
Thales CortAIx Labs
}

\begin{document}
\maketitle

\begin{abstract}
Achieving both high accuracy and topological continuity in road segmentation from satellite imagery is a critical goal for applications ranging from urban planning to disaster response. State-of-the-art methods often rely on Vision Transformers, which excel at capturing global context, yet their quadratic complexity is a significant barrier to efficient deployment, particularly for on-board processing in resource-constrained platforms. In contrast, emerging State Space Models like Mamba offer linear-time efficiency and are inherently suited to modeling long, continuous structures. We posit that these architectures have complementary strengths. To this end, we introduce PathMamba, a novel hybrid architecture that integrates Mamba's sequential modeling with the Transformer's global reasoning. Our design strategically uses Mamba blocks to trace the continuous nature of road networks, preserving topological structure, while integrating Transformer blocks to refine features with global context. This approach yields topologically superior segmentation maps without the prohibitive scaling costs of pure attention-based models. Our experiments on the DeepGlobe Road Extraction \cite{demir2018deepglobe} and Massachusetts Roads \cite{mnih2013machine} datasets demonstrate that PathMamba sets a new state-of-the-art. Notably, it significantly improves topological continuity, as measured by the APLS metric \cite{etten2019spacenet}, setting a new benchmark while remaining computationally competitive.
\end{abstract}

\section{Introduction}
\label{sec:intro}

\begin{figure}[t]
    \centering
    \begin{minipage}[c]{0.5\linewidth}
        \includegraphics[width=\linewidth]{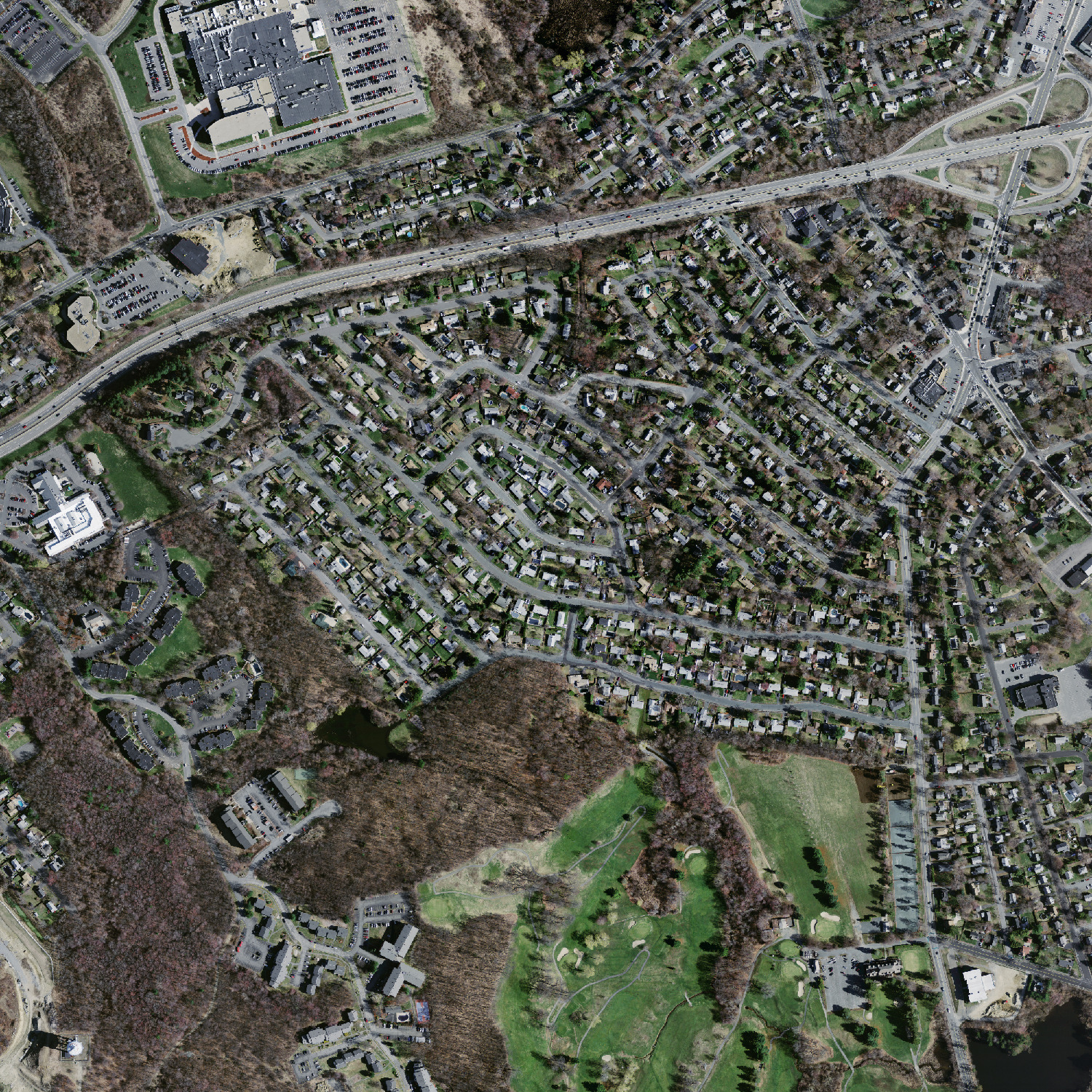}
    \end{minipage}%
    \begin{minipage}[c]{0.5\linewidth}
        \includegraphics[width=\linewidth]{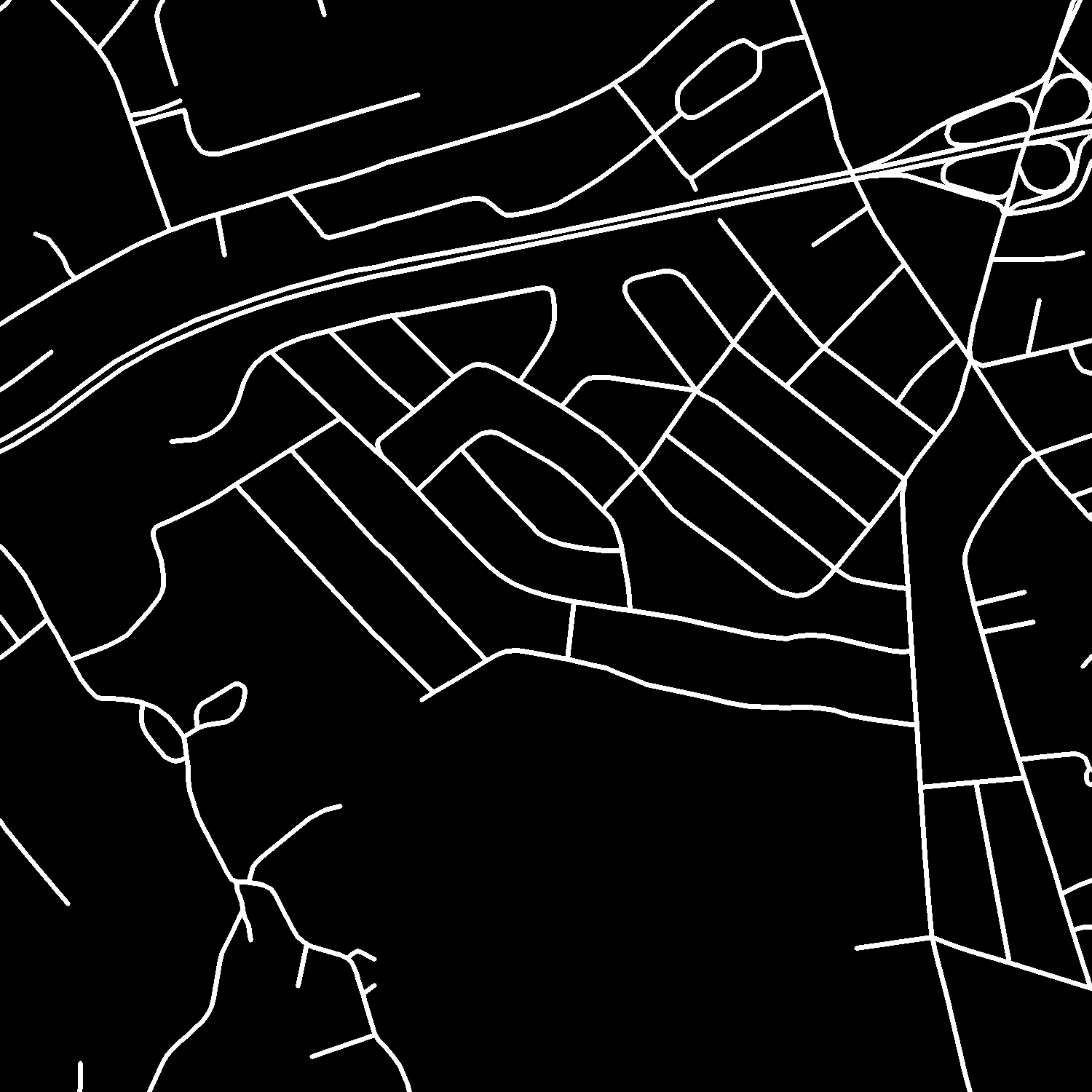}
    \end{minipage}
    \caption{An example from the Massachusetts Roads dataset illustrating the challenges of road segmentation. The input satellite image (left) contains significant occlusions from vegetation and complex shadows, while the goal is to produce a precise, topologically coherent ground-truth mask (right).}
    \label{fig:mass}
\end{figure}

Road segmentation from satellite imagery is a critical task in remote sensing with wide-ranging applications in autonomous navigation, urban planning, and disaster response \cite{mnih2010learning,mnih2013machine}. The primary goal is to produce a precise pixel-wise mask that delineates road networks \cite{long2015fully}. This task is inherently difficult due to numerous challenges, including frequent occlusions from buildings and vegetation, inconsistent lighting and shadows, the thin and elongated structure of roads, and severe class imbalance between road and background pixels \cite{cheng2017automatic, zhang2018road}. These factors, illustrated in \Cref{fig:mass}, make it especially hard to maintain the topological integrity of the predicted network, often resulting in fragmented segments \cite{mosinska2018beyond}.

For years, Convolutional Neural Networks (CNNs) \cite{shelhamer2016fully,ronneberger2015unet} have been the foundational architecture for segmentation models. Their strong inductive biases centered on locality are effective for general feature extraction \cite{nandakumar2023why}, but their intrinsically localized receptive fields struggle to model the long-range dependencies required to trace a road's path across an entire complex scene \cite{younesi2024comprehensive}. To overcome this limitation, Vision Transformers (ViTs) \cite{dosovitskiy2021image} were introduced, leveraging a self-attention mechanism to capture global context. This approach led to superior performance in many vision tasks, including road segmentation \cite{xie2021segformer,cao2021swinunet}. However, the quadratic complexity of self-attention with respect to input size makes ViTs computationally expensive and difficult to scale for high-resolution satellite imagery \cite{vaswani2017attention}.

Recently, State Space Models (SSMs) have emerged as a compelling alternative, challenging the dominance of Transformers \cite{gu2022efficiently}. Initially developed for natural language processing (NLP) to efficiently model long sequences, the Mamba architecture \cite{gu2024mamba} introduced a selective scanning mechanism that combines linear-time complexity with the ability to capture long-range dependencies. This success has spurred its adaptation to computer vision, leading to a new family of models. Architectures like Vision Mamba (ViM) \cite{zhu2024vision} and the hierarchical VMamba \cite{liu2024vmamba} have successfully adapted the sequential SSM core for 2D image processing by serializing patches into sequences, demonstrating performance on par with Transformers but with significantly better scalability.

Maintaining network connectivity is a critical challenge where standard pixel-wise losses often fall short \cite{mosinska2018beyond}. Current approaches to improve topology typically rely on two strategies: complex post-processing steps to reconnect fragmented road segments \cite{bastani2018roadtracer, mattyus2017deeproadmapper}, or the integration of specialized topological loss functions into the training process \cite{bandara2021spin, mosinska2018beyond}. Our work pushes this limit by proposing an architectural solution designed to inherently generate more topologically coherent segmentation maps, thereby reducing the reliance on such auxiliary techniques.

We posit that an optimal architecture for road segmentation can be achieved by combining the complementary strengths of Mamba and Transformers, a direction also explored in recent hybrid models \cite{hatamizadeh2025mambavision, zhang2025mhs}. Mamba's unique advantage lies in its selective state-space mechanism \cite{gu2024mamba}, which allows it to dynamically compress and carry relevant contextual information along a sequence. This is not merely a matter of efficiency; it is a fundamental architectural advantage for modeling long-range dependencies. Prior work on the Long Range Arena benchmark \cite{tay2020long} demonstrated that foundational SSMs could achieve near-perfect scores on the 
Path-X task, a challenge requiring a model to determine if two points are connected by a path of dashes amidst numerous distractor paths, where Transformers trained from scratch were unable to solve the task \cite{gu2022efficiently, amos2024never}. The conceptual parallel between this task and tracing occluded roads is striking: the path composed of dashes creates a visually discontinuous sequence, functionally equivalent to a road that is intermittently occluded by trees and shadows. In both scenarios, the core challenge is to maintain contextual memory and infer the underlying topological connection despite interruptions in the input data. This makes Mamba's state-space mechanism, which is designed to carry information across long sequences, exceptionally well-suited for modeling such structures. Conversely, the Transformer's strength is its content-based global self-attention \cite{vaswani2017attention}, which performs an all-pairs comparison of image patches. While computationally intensive, this mechanism is unparalleled for understanding complex, isotropic spatial relationships (like intersections) and resolving ambiguities by drawing context from the entire scene at once. We propose a novel hybrid backbone where Mamba blocks are primarily used to model the long, continuous structure of roads, while strategically placed Transformer blocks provide robust global reasoning to connect disparate segments. This targeted combination aims to preserve the topological continuity that Mamba excels at, while leveraging the Transformer's power for global scene understanding to improve overall accuracy, a core hypothesis that we validate quantitatively in our experimental analysis (see \Cref{subsec:val_hypothesis}).

Our main contributions are:
\begin{itemize}
    \item A novel hybrid Mamba-Transformer backbone that sequentially leverages Mamba's continuity modeling and the Transformer's global context aggregation within a single stage.
    \item State-of-the-art performance on the DeepGlobe Road Extraction benchmark across IoU, F1-score, and the topology-focused APLS metric \cite{etten2019spacenet}.
    \item The highest reported APLS score on the challenging Massachusetts Roads dataset, demonstrating superior topological integrity in the predicted road networks.
    \item Comprehensive ablation studies validating our architectural design choices and demonstrating the complementary contributions of the Mamba and Transformer components.
\end{itemize}

\section{Related Work}
\label{sec:related_work}

\paragraph{Pre-Deep Learning Methods.}
Before the widespread adoption of deep learning, road extraction from aerial and satellite imagery was tackled using a variety of classical computer vision and machine learning techniques \cite{wang2016review}. These methods often relied on handcrafted features and the explicit modeling of road properties. Prominent approaches included probabilistic and graph-based models, which were combined to identify and structure road networks \cite{unsalan2012road}. This family of methods included the use of geometric-stochastic models to find main roads \cite{barzohar1996automatic} and Gibbs point processes to handle complex road geometries \cite{stoica2004gibbs}. Complementary techniques focused on recovering line networks by detecting junctions \cite{chai2013recovering} or using active contour models, also known as snakes, to delineate road boundaries based on scale space analysis \cite{laptev2000automatic}. While foundational, these methods often struggled with the complex variations and occlusions present in real-world imagery, paving the way for data-driven deep learning solutions.

\paragraph{CNN-based Road Segmentation.}
Early and influential approaches to road segmentation relied on fully convolutional networks. Architectures like U-Net \cite{ronneberger2015unet} and DeepLab \cite{chen2017deeplab} set strong baselines. D-LinkNet \cite{zhou2018dlinknet} specifically tailored its design for road extraction by using dilated convolutions in its center path to enlarge the receptive field and preserve spatial information, achieving state-of-the-art results at the time. More recent works continue to build on these foundations, introducing refinements such as attention mechanisms and novel fusion strategies to further improve performance \cite{he2022road, wachtersimon2023}. However, the inherently local nature of convolutions remains a limitation for capturing global road network topology, a challenge that persists even in modern CNN designs \cite{abdollahi2021review}.

\paragraph{Transformer-based Models.}
Vision Transformers (ViTs) \cite{dosovitskiy2021image} and their hierarchical variants like the Swin Transformer \cite{liu2021swin} introduced global self-attention to vision, overcoming the limited receptive fields of CNNs \cite{zuo2022vision}. Models like SegFormer \cite{xie2021segformer} and Swin-Unet \cite{cao2021swinunet} extended this paradigm to semantic segmentation, demonstrating strong performance on road extraction benchmarks \cite{zhao2024road, jiang2022roadformer}. Despite their success, the quadratic complexity of self-attention poses a significant computational barrier for high-resolution remote sensing applications \cite{vaswani2017attention}.

\paragraph{State Space Models in Vision.}
State Space Models (SSMs) have recently been adapted for vision tasks as an efficient alternative to Transformers. Their theoretical strength in handling long sequences was empirically validated by precursors like S4 \cite{gu2022efficiently} on the demanding Long Range Arena benchmark \cite{tay2020long}. Notably, on the Path-X benchmark within this suite, which requires identifying a continuous logical path from a series of visual dashes, S4 achieved a high score while Transformer-based models trained from scratch failed. The task's use of a dashed path serves as a strong proxy for real-world continuity challenges, such as tracing road networks fragmented by occlusions. This result highlights that the core SSM mechanism is inherently superior for tasks demanding long-range spatial continuity. Mamba \cite{gu2024mamba} builds upon this success by introducing a selective scan mechanism that further enhances the ability to model these dependencies with linear-time complexity. This has led to vision-specific adaptations like Vision Mamba (ViM) \cite{zhu2024vision} and the hierarchical VMamba \cite{liu2024vmamba}. VMamba, in particular, uses a Cross-Scan Module to traverse image patches in multiple directions, effectively capturing 2D spatial context by applying the sequential SSM mechanism along these generated paths. Concurrently, other works have explored Mamba's potential specifically for remote sensing. For instance, RS-Mamba \cite{zhao2024rsmamba} proposed a specialized architecture that also leverages Mamba for segmentation and has shown competitive results. Hybrid architectures like MambaVision \cite{hatamizadeh2025mambavision} have also emerged, combining Mamba and Transformer blocks to leverage their complementary strengths. Our work builds on this direction but proposes a specific sequential arrangement tailored for the unique challenges of road segmentation.

\paragraph{Approaches for Topological Coherence.}
Pixel-wise metrics like IoU often fail to penalize topological errors, such as disconnected road segments, which are critical for navigation applications. This has motivated research into methods that explicitly optimize for topology \cite{etten2019spacenet}. These approaches typically fall into two categories: complex post-processing steps to reconnect fragmented segments \cite{bastani2018roadtracer}, or the integration of specialized topological loss functions into the training process \cite{mosinska2018beyond}. Architectural innovations, such as the graph-based reasoning in SPIN \cite{bandara2021spin}, often function as refinement modules that operate on features from a standard backbone to enforce connectivity. While effective, these methods are primarily reactive, designed to repair topological errors after they occur. Our work diverges from this paradigm by proposing a preventative, architectural solution. We hypothesize that by integrating a mechanism inherently suited to modeling continuity, the Mamba SSM, directly into the backbone, the network can learn to generate feature representations that are already topologically sound, thereby reducing the dependency on complex loss functions or post-processing heuristics.

\section{Methodology}
\label{sec:method}

\subsection{State Space Model Preliminaries}
At its core, PathMamba is built upon the State Space Model (SSM) framework. This section briefly reviews its foundational principles, as detailed in the original Mamba paper \cite{gu2024mamba} and adapted for vision in architectures like VMamba \cite{liu2024vmamba}. A continuous-time SSM maps a 1D input signal \( x(t) \in \mathbb{R} \) to an output \( y(t) \in \mathbb{R} \) via a latent state \( h(t) \in \mathbb{R}^N \). The system is governed by a set of linear ordinary differential equations:
\begin{align}
    h'(t) &= A h(t) + B x(t) \\
    y(t) &= C h(t) + D x(t)
\end{align}
where \( A \in \mathbb{R}^{N \times N} \) is the state matrix and \( B \in \mathbb{R}^{N \times 1} \), \( C \in \mathbb{R}^{1 \times N} \), \( D \in \mathbb{R} \) are projection matrices.

A crucial characteristic of modern SSMs, including Mamba, is that they operate as single-input single-output (SISO) systems. To handle the multi-channel feature maps common in computer vision, where an input can be considered \( x(t) \in \mathbb{R}^D \), the SSM processes each of the \( D \) channels independently. This results in a much larger effective state size of \( N \times D \), where \( N \) is the state expansion factor. This expansion of the latent state is a key element that allows the model to capture complex relationships within information-dense data like images \cite{gu2024mamba}. 

To be used in deep learning, this system must be discretized. Using a timestep \( \Delta \), the continuous parameters are transformed into discrete counterparts \( \bar{A}, \bar{B} \) via a discretization rule like the Zero-Order Hold (ZOH):
\begin{align}
    \bar{A} &= \exp(\Delta A) \\
    \bar{B} &= (\Delta A)^{-1}(\exp(\Delta A) - I) B
\end{align}
The discrete-time SSM is then given by:
\begin{align}
    h_t &= \bar{A} h_{t-1} + \bar{B} x_t \\
   y_t &= \bar{C} h_ti
\end{align}

Following the convention in recent works \cite{gu2024mamba, liu2024vmamba}, we omit the direct feedthrough matrix \( D \) as its function is captured by residual connections in the overall block architecture. 

Mamba enhances this formulation with a selective scan mechanism, where the \( B, C \) matrices and the timestep \( \Delta \) are dynamically generated from the input data itself. This makes the SSM input-dependent and time-variant, allowing the model to selectively focus on or ignore parts of the input sequence. This selectivity breaks the property of time-invariance, which precludes the use of standard convolution. Mamba therefore employs an efficient parallel scan algorithm for computation instead \cite{gu2024mamba}.

\subsection{Architectural Foundations}

Our architecture is built upon the hierarchical design popularized by Swin Transformer \cite{liu2021swin} and VMamba \cite{liu2024vmamba}. The model processes an input image through four stages, progressively downsampling the spatial resolution while increasing the channel dimension. This multi-scale feature extraction is crucial for semantic segmentation, as it allows the model to capture both fine-grained details and high-level contextual information simultaneously \cite{zhao2017pyramid, lin2017feature, chen2017deeplab}.

\begin{figure}[t]
  \centering
  \includegraphics[width=0.62\linewidth]{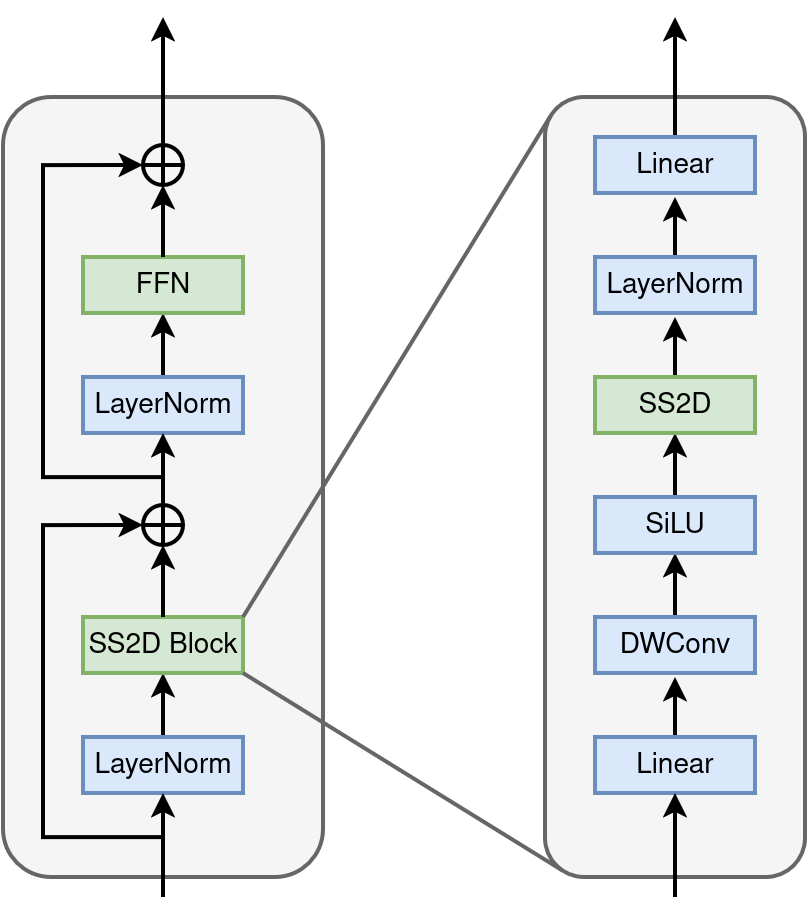}
  \caption{Architecture of a Visual State Space (VSS) block, which forms the basis of our Mamba stages. It uses an SSM for token-mixing and an FFN for channel-mixing.}
  \label{fig:vss}
\end{figure}

The core component in the Mamba-based stages is the Visual State Space (VSS) block, illustrated in \Cref{fig:vss}. It follows a MetaFormer structure \cite{yu2022metaformer}, which separates token-mixing from channel-mixing. The token-mixing is performed by a 2D-aware SSM module, while a standard Feed-Forward Network (FFN) handles channel-mixing. To apply the inherently sequential SSM to 2D image data, we employ the cross-scan strategy from VMamba. This involves serializing the image patches by scanning them along four cardinal directions, allowing the SSM to capture comprehensive spatial context.

\subsection{Hybrid Mamba-Transformer Backbone}

The primary contribution of this work is a novel hybrid backbone that strategically combines Mamba and Transformer blocks. Our design is motivated by the hypothesis that Mamba and Transformers have complementary strengths for road segmentation. This hypothesis is directly informed by prior work on long-range benchmarks. The task of road segmentation, especially with occlusions, is functionally analogous to the Path-X benchmark \cite{tay2020long}, where the challenge is to identify a logically continuous path composed of visually discontinuous dashes. These interruptions are a direct proxy for the challenge of tracing a road network that is intermittently hidden by occlusions like trees and shadows. Given that SSMs have proven uniquely capable of solving this task while Transformers trained from scratch have failed \cite{gu2022efficiently}, we posit that Mamba's architecture is fundamentally better suited for establishing topological continuity. As confirmed in our analysis in \Cref{subsec:val_hypothesis}, Mamba excels at modeling these long, continuous structures. In contrast, Transformers excel at global contextual reasoning, making them powerful for improving overall classification accuracy. Our proposed hybrid stage is therefore designed to harness both capabilities.

\begin{figure*}[t]
  \centering
  \includegraphics[width=\linewidth]{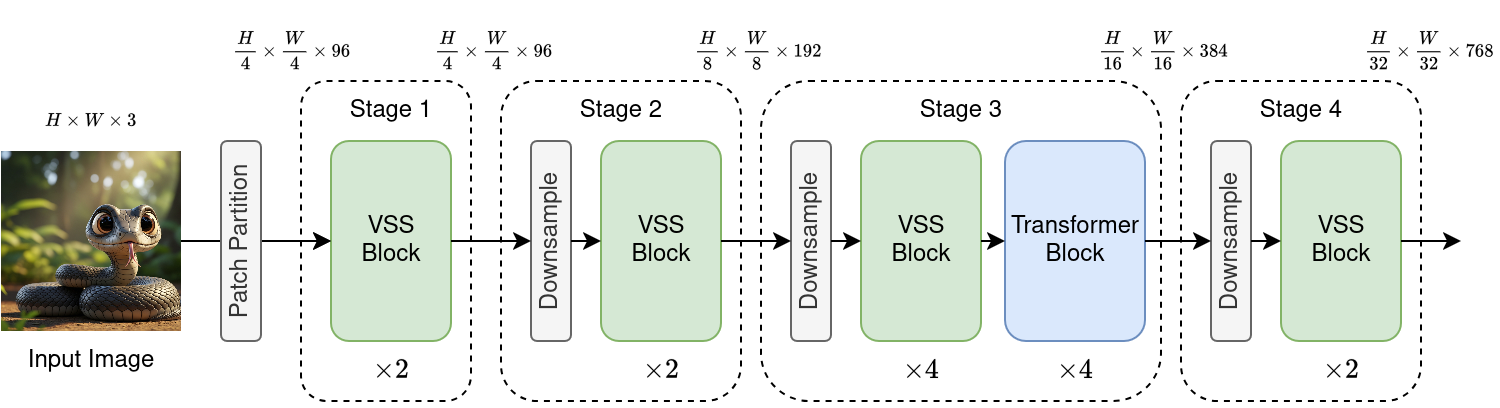}
  \caption{The architecture of our proposed hybrid Mamba-Transformer backbone. Stages 1, 2, and 4 consist of VSS (Mamba) blocks. Stage 3 is a hybrid stage containing a sequence of VSS blocks followed by standard Transformer (Attention) blocks to integrate continuity modeling with global context aggregation. All stages are connected to a UperNet decoder head.}
  \label{fig:architecture}
\end{figure*}

We propose a four-stage hierarchical backbone, as shown in \Cref{fig:architecture}. The first, second, and fourth stages are composed entirely of VSS blocks. The key innovation lies in the \textbf{third stage}, which is a hybrid design. It begins with a series of VSS blocks to efficiently process features and model continuity, followed by a series of standard Transformer blocks (Multi-Head Self-Attention).

The sequential arrangement within our hybrid stage, VSS blocks followed by Transformer blocks, is a deliberate design choice. The initial Mamba blocks act as efficient continuity modelers, tracing the elongated structure of roads and propagating contextual information along their paths. This process generates feature maps that are already imbued with a strong sense of topological structure. The subsequent Transformer blocks then operate on these semantically enriched features. Their global self-attention mechanism is perfectly suited to resolve complex spatial ambiguities, such as multi-road intersections, and to integrate information from disconnected regions of the image, effectively refining the continuous paths identified by Mamba.

This arrangement allows the model to first capture continuous features efficiently with Mamba at a medium resolution. Then, the Transformer blocks operate on these feature maps to perform global context aggregation. Placing the computationally intensive attention mechanism in a deeper stage, where the spatial resolution is reduced, makes it computationally tractable while maximizing its impact.

\subsection{Segmentation Head}
To generate the final segmentation mask, we use the UperNet decoder head \cite{xiao2018unified} with all our backbone experiments. UperNet effectively fuses multi-scale features from the backbone's four stages using a combination of a Pyramid Pooling Module (PPM) \cite{zhao2017pyramid} and a Feature Pyramid Network (FPN) \cite{lin2017feature}. This creates a feature representation that is rich in both high-level semantic context and low-level spatial detail. Using a consistent, powerful decoder allows us to focus our analysis on the performance of the encoder backbones.

\section{Experiments}
\label{sec:experiments}

\subsection{Experimental Setup}

\paragraph{Datasets.}
We evaluate our model on two public benchmarks:
\begin{itemize}
    \item \textbf{DeepGlobe Road Extraction Dataset} \cite{demir2018deepglobe}: Contains 6,226 satellite images (1024$\times$1024) with a Ground Sampling Distance (GSD) of 0.5 m, covering diverse geographic regions.
    \item \textbf{Massachusetts Roads Dataset} \cite{mnih2013machine}: Comprises 1,171 aerial images (1500$\times$1500) with a GSD of 1 m, covering urban and suburban areas. A key challenge is its annotation style, where all roads are rasterized to a fixed 7-pixel width.
\end{itemize}

\paragraph{Evaluation Metrics.}  
We use standard pixel-wise metrics for the road class: Intersection over Union (IoU) and F1-Score. To specifically assess the topological integrity of the road network, we use the Average Path Length Similarity (APLS) metric \cite{etten2019spacenet}. To compute this metric, the binary segmentation masks (both predicted and ground-truth) are first converted into graph structures via morphological skeletonization. APLS compares shortest path lengths in these graphs, yielding a score from 0 (dissimilar) to 1 (identical). It is defined as:
\begin{align*}
    \text{APLS} = 1 - \frac{1}{N} \sum \min \left\{1,\ \frac{|L(a,b) - L(a',b')|}{L(a,b)} \right\}
\end{align*}
where \( N \) is the number of paths in the ground truth graph, \( L(a,b) \) is a path length in the ground truth, and \( L(a',b') \) is the corresponding path length in the prediction.

\paragraph{Implementation Details.} 
Our model and all baselines are implemented using the MMSegmentation v1.2.2 framework. For fair comparison, all models benchmarked against the state-of-the-art are initialized with weights pretrained on ImageNet-1K and subsequently on ADE20K. The ADE20K benchmark was run following its standard training protocol. All ablation studies were trained from scratch on the DeepGlobe dataset. We use the AdamW optimizer with a learning rate of $6 \times 10^{-5}$ and a weight decay of 0.01. The loss function is a combination of binary cross-entropy and Dice loss, a hybrid approach effective for balancing pixel-wise supervision with robustness to class imbalance \cite{xu2023comparativea}. The learning rate follows a linear warm-up for 1,500 iterations, followed by a polynomial decay schedule. Models were trained for 160k iterations on DeepGlobe and 80k iterations on the smaller Massachusetts Roads dataset to prevent overfitting. Experiments were conducted on NVIDIA A100 GPUs.

\paragraph{Baselines.}
We compare our model against similarly-sized, state-of-the-art architectures. Our selected baselines include: a strong CNN in DeepLabv3+ \cite{chen2017deeplab} with a ResNet-101 backbone; leading Transformer models like SegFormer \cite{xie2021segformer} with an MIT-B3 backbone and Swin-UperNet \cite{liu2021swin} (Swin-T backbone); and recent Mamba-based architectures such as VMamba-T \cite{liu2024vmamba} and MambaVision-T \cite{hatamizadeh2025mambavision}, both using the UperNet decoder. This selection provides a comprehensive comparison across different architectural paradigms.

\subsection{Generalization Benchmarks}
\label{sec:benchmarks}

To first establish the general feature extraction capability of our backbone, we benchmark it on the ImageNet-1K classification task \cite{deng2009imagenet}. As shown in \Cref{tab:imagenet}, PathMamba achieves a Top-1 accuracy of \textbf{83.1\%}, outperforming other leading architectures of a similar scale. This demonstrates the fundamental strength of our hybrid design.

\begin{table}[h]
\centering
\caption{Performance on the \textbf{ImageNet-1K} validation set. Best in \textbf{bold}, second in \underline{underline}.}
\label{tab:imagenet}
\resizebox{0.9\columnwidth}{!}{%
\begin{tabular}{lcccc}
\toprule
Method & Backbone & Params (M) & Top-1 Acc. (\%) \\
\midrule
Swin \cite{liu2021swin} & Swin-T & 29 & 81.3 \\
VMamba \cite{liu2024vmamba} & VMamba-T & 30 & 82.6 \\
MambaVision \cite{hatamizadeh2025mambavision} & MambaVision-T & 35 & \underline{82.7} \\
\textbf{Ours} & PathMamba & 31 & \textbf{83.1} \\
\bottomrule
\end{tabular}%
}
\end{table}

We further validate its versatility for dense prediction tasks on the challenging ADE20K dataset \cite{zhou2017scene}. As shown in \Cref{tab:ade20k}, our model achieves a new state-of-the-art mIoU of \textbf{48.6\%}. This result surpasses leading Mamba and Transformer-based architectures, demonstrating that our hybrid design is a powerful and versatile feature extractor effective for general-purpose segmentation beyond remote sensing.

\begin{table}[h]
\centering
\caption{Performance on the \textbf{ADE20K} validation set (single-scale mIoU). Our model is compared against other architectures of a similar scale. Best in \textbf{bold}, second in \underline{underline}.}
\label{tab:ade20k}
\resizebox{\columnwidth}{!}{%
\begin{tabular}{lcccc}
\toprule
Method & Backbone & Params (M) & FLOPs (G) & mIoU (\%) \\
\midrule
Swin-UperNet \cite{liu2021swin} & Swin-T & 59.0 & 940 & 44.4 \\
DeepLabV3+ \cite{chen2017deeplab} & R-101-D8 & 60.1 & 1017 & 45.5 \\
MambaVision \cite{hatamizadeh2025mambavision} & MambaVision-T & 61.7 & 933 & 46.0 \\
SegFormer \cite{xie2021segformer} & MIT-B3 & 44.6 & 239 & 47.8 \\
VMamba \cite{liu2024vmamba} & VMamba-T & 60.9 & 942 & \underline{47.9} \\
\textbf{Ours} & PathMamba & 61.7 & 946 & \textbf{48.6} \\
\bottomrule
\end{tabular}%
}
\end{table}

\subsection{Comparison with State-of-the-Art}

\begin{table}[t]
\centering
\caption{Results on the \textbf{Massachusetts Roads} test set. Our model achieves the best topological score (APLS) while remaining competitive on pixel-wise metrics and inference speed. Best in \textbf{bold}, second in \underline{underline}.}
\label{tab:sota_massachusetts}
\resizebox{\columnwidth}{!}{%
\begin{tabular}{lcccccc}
\toprule
Method & IoU (\%) & F1 (\%) & APLS & Params (M) & FLOPs (G) & FPS \\
\midrule
SegFormer \cite{xie2021segformer}      & 66.94 & 80.20 & 78.40 & 44.6 & 796 & 6.17 \\
MambaVision \cite{hatamizadeh2025mambavision}    & 67.31 & 80.46 & 78.00 & 61.7 & 2188 & 9.67 \\
DeepLabv3+ \cite{chen2017deeplab}     & 67.51 & 80.61 & 78.48 & 60.2 & 2288 & 10.2 \\
Vanilla-VMamba \cite{liu2024vmamba} & 67.76 & 80.78 & 78.32 & 53.5 & 2058 & 5.22 \\
Swin-UperNet \cite{liu2021swin}   & 68.20 & 81.10 & 79.53 & 58.9 & 2098 & 11.08 \\
VMamba \cite{liu2024vmamba}         & \textbf{68.26} & \textbf{81.14} & \underline{79.66} & 61.8 & 2111 & \textbf{13.8} \\
\textbf{Ours}  & \underline{68.20} & \underline{81.09} & \textbf{79.88} & 61.8 & 2119 & \underline{9.22} \\
\bottomrule
\end{tabular}
}
\end{table}

\begin{table}[t]
\centering
\caption{Results on the \textbf{DeepGlobe Road Extraction} test set. Our hybrid model sets a new state-of-the-art across all metrics.}
\label{tab:sota_deepglobe}
\resizebox{\columnwidth}{!}{%
\begin{tabular}{lcccccc}
\toprule
Method & IoU (\%) & F1 (\%) & APLS (\%) & Params (M) & FLOPs (G) & FPS \\
\midrule
Swin-UperNet \cite{liu2021swin} & 70.29 & 82.55 & 75.74 & 59.0 & 940 & 22.8 \\
DeepLabv3+ \cite{chen2017deeplab} & 70.35 & 82.59 & 75.38 & 60.1 & 1017 & 23.2 \\
MambaVision \cite{hatamizadeh2025mambavision} & 70.55 & 82.73 & 77.11 & 61.7 & 933 & 25.2 \\
Vanilla-VMamba \cite{liu2024vmamba} & 71.67 & 83.50 & 79.72 & 53.5 & 919 & 15.0 \\
SegFormer \cite{xie2021segformer} & 71.94 & 83.68 & 78.82 & 44.6 & 239 & 21.2 \\
VMamba \cite{liu2024vmamba} & \underline{72.08} & \underline{83.78} & \underline{79.42} & 60.9 & 942 & \textbf{26.8} \\
\textbf{Ours} & \textbf{72.19} & \textbf{83.85} & \textbf{80.03} & 61.7 & 946 & \underline{22.9} \\
\bottomrule
\end{tabular}
}
\end{table}

As shown in \Cref{tab:sota_massachusetts} and \Cref{tab:sota_deepglobe}, our hybrid architecture achieves outstanding performance.

On \textbf{Massachusetts Roads}, our model obtains the highest APLS score of \textbf{79.88\%}, demonstrating its superior ability to generate topologically correct road networks. This is particularly significant given the dataset's fixed-width annotation artifact, which makes APLS a more reliable metric than IoU or F1-score for evaluating functional quality.

On \textbf{DeepGlobe Road Extraction}, our model establishes a new state-of-the-art, outperforming all baselines across all metrics. It achieves an IoU of \textbf{72.19\%}, an F1-score of \textbf{83.85\%}, and a top APLS score of \textbf{80.03\%}. This consistent superiority confirms our hypothesis that combining Mamba's continuity modeling with the Transformer's global reasoning leads to more accurate and complete segmentations, all while maintaining a comparable computational footprint to other leading models.

\subsection{Ablation Studies}
We conducted extensive ablation studies on the DeepGlobe dataset to validate our design choices. For computational efficiency and a clear assessment of architectural changes, all models in this section were trained from scratch. As expected, this protocol results in performance metrics that are lower than their fully pretrained counterparts (e.g., our model achieves 69.10\% IoU from scratch vs. 72.19\% with pretraining). However, the relative performance differences between configurations provide a valid basis for our conclusions.

\paragraph{Impact of Hybrid Stage Arrangement.}
Our investigation into the arrangement of Mamba ('m') and attention ('a') blocks, detailed in \Cref{tab:hybrid_ablation}, reveals a nuanced trade-off between pixel-wise accuracy (IoU/F1) and topological continuity (APLS). No single configuration achieved superiority across all metrics. Instead, our proposed `mmmm-aaaa` design and the alternating `ma-ma-ma-ma` configuration emerge as the two top-performing models, forming a Pareto front.

Specifically, our `mmmm-aaaa` architecture achieves the best performance on the primary segmentation metrics, with an IoU of \textbf{69.10\%} and an F1-score of \textbf{81.72\%}. The `ma-ma-ma-ma` configuration, while performing slightly worse on these metrics, secures a marginal lead in its APLS score. Given that the gains our model achieves in IoU and F1-score are more significant than its minor deficit in APLS, our proposed architecture offers the most compelling and well-balanced performance profile. We therefore select the `mmmm-aaaa` configuration as our final design.

\begin{table}[h]
\centering
\caption{Ablation on the arrangement of Mamba ('m') and attention ('a') blocks in the hybrid stage. Our proposed `mmmm-aaaa` configuration achieves the best performance on pixel-wise metrics (IoU/F1).}
\label{tab:hybrid_ablation}
\resizebox{0.9\columnwidth}{!}{%
\begin{tabular}{lccc}
\toprule
Hybrid Stage Config. & IoU (\%) & F1 (\%) & APLS (\%) \\
\midrule
\textbf{mmmm-aaaa (Ours)} & \textbf{69.10} & \textbf{81.72} & 75.33 \\
ma-ma-ma-ma & 68.84 & 81.54 & \textbf{75.44} \\
am-am-am-am & 68.88 & 81.57 & 75.02 \\
aaaa-mmmm & 68.85 & 81.55 & 74.64 \\
mmmmmmm-a & 68.64 & 81.40 & 74.79 \\
\bottomrule
\end{tabular}%
}
\end{table}

\paragraph{Necessity of the SSM Component.}
To verify the importance of Mamba's core SSM mechanism, especially in light of recent work questioning its necessity for all vision tasks \cite{yu2024mambaout}, we replaced the SSM component in the VSS blocks with a simple identity mapping. As shown in \Cref{tab:ssm_ablation}, removing the SSM from any part of the network leads to a significant drop in performance. Removing it entirely results in a catastrophic performance collapse, with the APLS score dropping by over 7 points. This definitively demonstrates that the selective state space mechanism is critical for achieving high-quality topological results in this domain.

\begin{table}[h]
\centering
\caption{Ablation on the removal of the SSM component from different stages. The SSM is critical for high performance.}
\label{tab:ssm_ablation}
\resizebox{0.9\columnwidth}{!}{%
\begin{tabular}{lccc}
\toprule
SSM Removed from Stages & IoU (\%) & F1 (\%) & APLS (\%) \\
\midrule
- (Baseline) & \textbf{69.10} & \textbf{81.72} & \textbf{75.33} \\
Stage 1, 2 & 68.72 & 81.46 & 74.63 \\
All Stages & 66.98 & 80.22 & 68.34 \\
\bottomrule
\end{tabular}%
}
\end{table}

\paragraph{Impact of Scan Strategy.}
To apply the sequential SSM to 2D images, a scanning strategy is required to serialize the image patches. Our architecture adopts the cross-scan method from VMamba \cite{liu2024vmamba} as its default. To validate this choice, we compared it against a range of alternative scanning patterns from recent literature, including uni/bi-directional scans \cite{liu2024vmamba}, vertical-2D scan \cite{zhang20252dmamba}, omni-directional scan \cite{zhao2024rsmamba}, local scan \cite{huang2024localmamba}, and fractal scan \cite{tang2024scalable}. As shown in \Cref{tab:scan_ablation}, while cross-scan achieves the best overall performance, the differences between the top methods are minor. This suggests that our model is largely robust to the specific choice of scanning path, a finding that aligns with other recent studies questioning the criticality of complex scan paths for vision tasks \cite{zhu2024rethinking}. This indicates that the core hierarchical and hybrid nature of our architecture is the primary driver of its strong performance.

\begin{table}[h]
\centering
\caption{Ablation on different 2D scan strategies. The Cross-Scan method used in our model provides the best overall performance.}
\label{tab:scan_ablation}
\resizebox{\columnwidth}{!}{%
\begin{tabular}{lccc}
\toprule
Scan Strategy & IoU (\%) & F1 (\%) & APLS (\%) \\
\midrule
\textbf{Cross-Scan (Ours)} \cite{liu2024vmamba} & \textbf{69.10} & \textbf{81.72} & \textbf{75.33} \\
Vertical-2D \cite{zhang20252dmamba} & 68.94 & 81.61 & 75.18 \\
Local-Scan \cite{huang2024localmamba} & 68.71 & 81.45 & 74.69 \\
Omni-Scan \cite{zhao2024rsmamba} & 68.77 & 81.50 & 74.66 \\
Bi-directional \cite{liu2024vmamba} & 69.00 & 81.66 & 74.62 \\
Uni-directional \cite{liu2024vmamba} & 68.79 & 81.51 & 73.94 \\
Fractal-Scan \cite{tang2024scalable} & 68.41 & 81.24 & 73.78 \\
\bottomrule
\end{tabular}%
}
\end{table}

\paragraph{On the Importance of Pretraining for Topology.}
The performance gap between our models trained from scratch and their pretrained counterparts is most pronounced on the APLS metric. This aligns with recent findings that long-sequence models benefit immensely from data-driven priors acquired during large-scale pretraining \cite{amos2024never}. In the context of road segmentation, these priors provide a rich, general understanding of common visual concepts like trees and buildings. A model trained from scratch may learn to associate roads with simple textural features, causing it to terminate a segment when faced with an occlusion. In contrast, a pretrained model can leverage its world knowledge to recognize the occluding object and infer the road's continuous path through the interruption. This ability is paramount for maintaining the topological integrity measured by APLS.

\subsection{Validating the Architectural Hypothesis}
\label{subsec:val_hypothesis}
To validate our foundational hypothesis regarding the complementary strengths of Mamba and Transformers, we compared different backbone configurations at the stage level. We benchmarked a pure Mamba backbone against hybrids where entire Mamba stages were replaced by Transformer stages. The results, summarized in \Cref{tab:hypothesis_validation}, reveal a distinct trade-off between pixel-level and topology-level performance.

\begin{table}[h]
\centering
\caption{Analysis of architectural contributions at the stage level on the DeepGlobe dataset. Each letter represents a full stage in the four-stage backbone, composed entirely of either Mamba ('m') or Transformer ('a') blocks. Best values are in \textbf{bold}.}
\label{tab:hypothesis_validation}
\resizebox{\columnwidth}{!}{%
\begin{tabular}{lcccc}
\toprule
\textbf{Backbone Stage Configuration} & \textbf{IoU (\%)} & \textbf{F1 (\%)} & \textbf{APLS (\%)} \\
\midrule
m $\rightarrow$ m $\rightarrow$ m $\rightarrow$ m (All-Mamba)      & 68.25          & 81.13          & \textbf{74.17} \\
m $\rightarrow$ m $\rightarrow$ m $\rightarrow$ a                & 67.88          & 80.87          & 73.94 \\
m $\rightarrow$ m $\rightarrow$ a $\rightarrow$ m                & 68.90          & 81.58          & 73.68 \\
\textbf{m $\rightarrow$ m $\rightarrow$ a $\rightarrow$ a} (Late-Attention) & \textbf{69.16} & \textbf{81.77} & 71.90 \\
\bottomrule
\end{tabular}%
}
\end{table}

The results provide strong quantitative support for our hypothesis. The pure Mamba backbone (m $\rightarrow$ m $\rightarrow$ m $\rightarrow$ m) achieves the highest APLS score, confirming its proficiency in modeling the continuous nature of road networks. Conversely, replacing later Mamba stages with Transformer stages (m $\rightarrow$ m $\rightarrow$ a $\rightarrow$ a) boosts the IoU score, demonstrating the Transformer's superior capability for global context aggregation, which aids in overall pixel classification. However, this gain in pixel-wise accuracy comes at the cost of a significant drop in topological continuity. This trade-off motivates our final architecture, which integrates these components within a single stage to achieve a more optimal balance.

\subsection{Qualitative Analysis}
\begin{figure}[h!]
    \centering
    \includegraphics[width=\columnwidth]{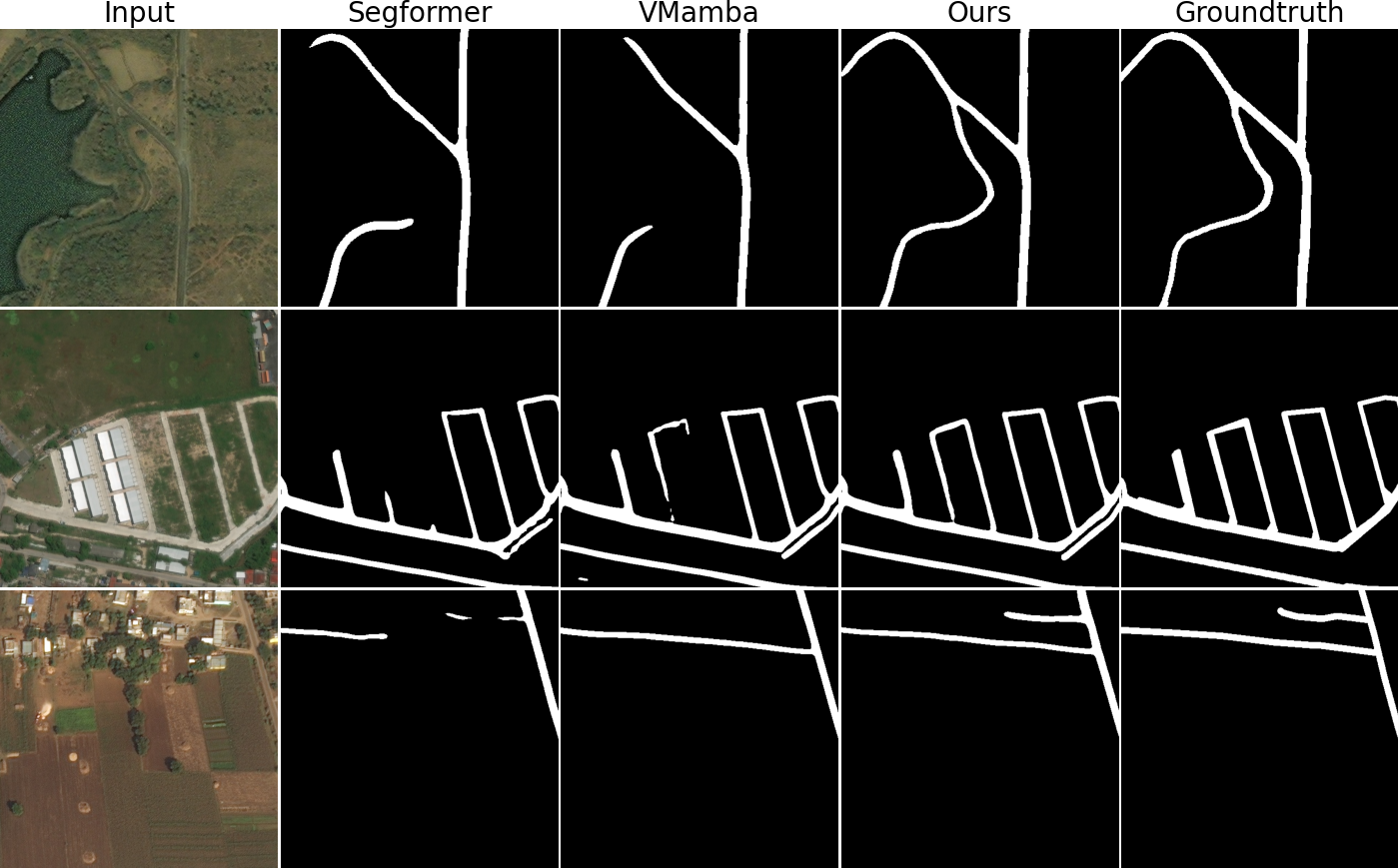}
    \caption{Qualitative comparison of our model against Segformer and VMamba. In the second row, both baselines produce disconnected masks, whereas our model correctly identifies the continuous road network. In the third row, our model accurately segments fine-grained details under tree occlusion, which are missed by the other methods.}
    \label{fig:qualitative_analysis}
\end{figure}

\begin{figure}[h!]
    \centering
    \begin{subfigure}{0.32\columnwidth}
        \includegraphics[width=\linewidth]{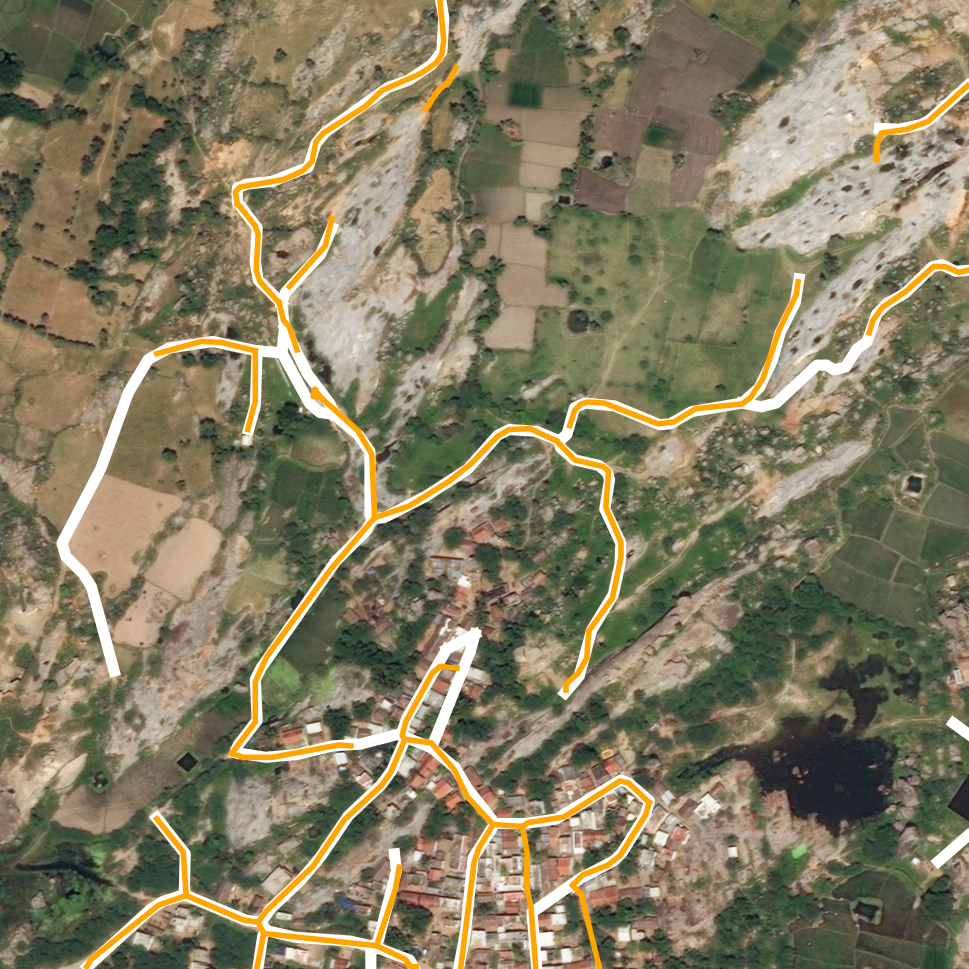}
        \caption{SegFormer \\ (APLS: 29.80\%)}
    \end{subfigure}
    \hfill
    \begin{subfigure}{0.32\columnwidth}
        \includegraphics[width=\linewidth]{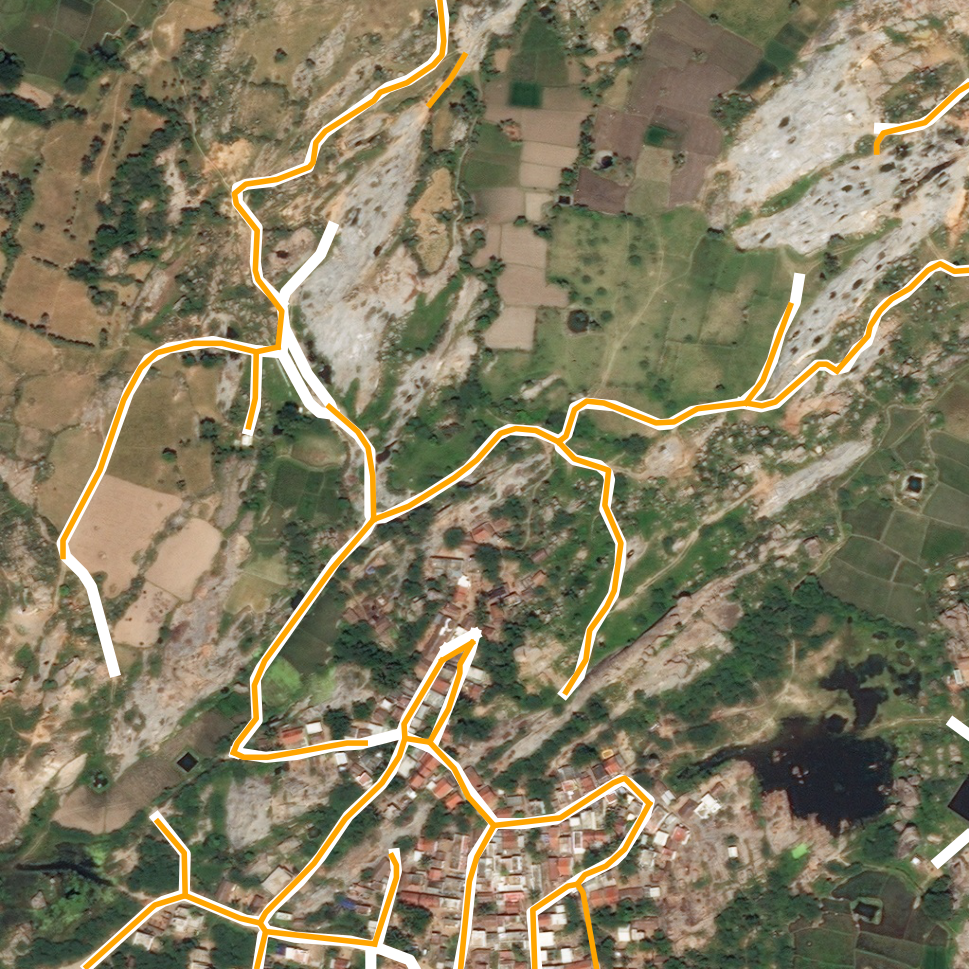}
        \caption{VMamba \\ (APLS: 42.20\%)}
    \end{subfigure}
    \hfill
    \begin{subfigure}{0.32\columnwidth}
        \includegraphics[width=\linewidth]{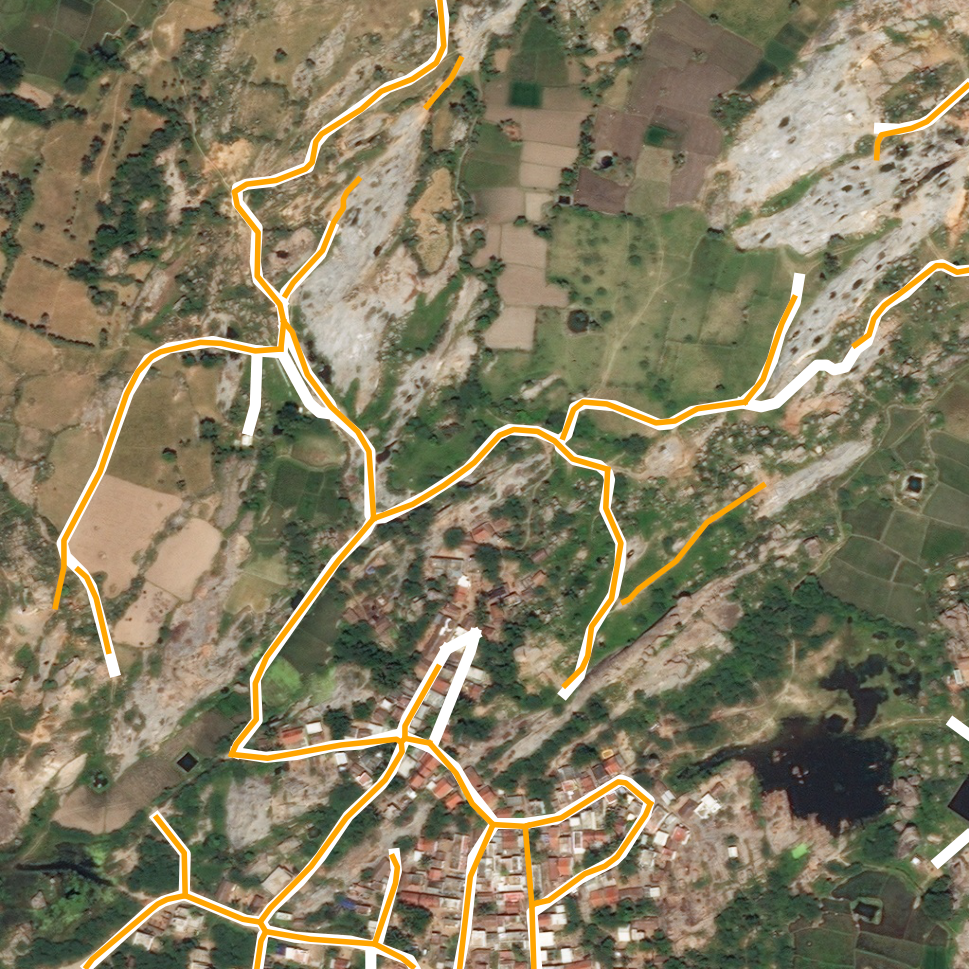}
        \caption{Ours \\ (APLS: 70.00\%)}
    \end{subfigure}
    \caption{A challenging sample featuring a rural road with inconsistent texture. Our model (c) maintains connectivity where baselines (a, b) fail due to surface variability, resulting in a significantly higher APLS score. (White = Ground Truth, Orange = Prediction).}
    \label{fig:qualitative_comp}
\end{figure}

To visually substantiate our quantitative results, we present a qualitative analysis in \Cref{fig:qualitative_analysis} and \Cref{fig:qualitative_comp}. These figures illustrate our model's superior ability to preserve road network integrity in challenging real-world scenarios.

\Cref{fig:qualitative_analysis} showcases several examples where our hybrid model excels compared to strong baselines like SegFormer and VMamba. Across different scenes, our model demonstrates a clear advantage in generating complete and continuous road masks, successfully avoiding the fragmentation that plagues the other methods. This is particularly evident in cases with complex intersections or where fine-grained road details are present.

\Cref{fig:qualitative_comp} focuses on a difficult case with a rural road of inconsistent texture, which often causes standard models to lose track of the path. While both SegFormer and VMamba produce fractured predictions, our model successfully maintains a continuous path. This visual success is directly quantified by the accompanying APLS scores, where our model achieves a dramatic improvement over the baselines, reinforcing the conclusion that our hybrid design leads to more reliable road segmentation.

\section{Conclusion}
\label{sec:conclusion}
In this paper, we introduced PathMamba, a novel hybrid Mamba-Transformer architecture for road segmentation from satellite imagery. By strategically combining the linear-time efficiency of Mamba for modeling continuous features with the powerful global reasoning of Transformers, our model leverages their combined strengths to yield superior performance.

Our experiments demonstrate that this hybrid design sets a new state-of-the-art on the DeepGlobe benchmark and, most importantly, achieves the highest topological accuracy (APLS score) on both the DeepGlobe Road Extraction and Massachusetts Roads datasets. This confirms that our architecture produces more coherent and functionally useful road networks. Comprehensive ablation studies validated our specific design choice of using Mamba blocks to first model continuity before refining features with self-attention.

Future work could explore scaling the architecture, investigating more advanced Mamba modules, and focusing on deployment-oriented optimizations like quantization to further enhance its suitability for real-world, resource-constrained applications in remote sensing.

{
    \small
    \bibliographystyle{ieeenat_fullname}
    \bibliography{main}
}

\clearpage
\setcounter{page}{1}
\maketitlesupplementary

\section{Reproducibility Details}
\label{sec:reproducibility}

This supplementary section provides additional details regarding our experimental setup to ensure the reproducibility of our results. We cover dataset splits, data preprocessing and augmentation, and the specific hyperparameter configuration for our main model. Our implementation is built upon the MMSegmentation v1.2.2 framework. We intend to release our code and pretrained models to facilitate further research.

\subsection{Datasets and Preprocessing}
We used standard, publicly available splits for all datasets to ensure fair and consistent comparison.

\paragraph{DeepGlobe Road Extraction.} We followed the official split from the CVPR 2018 challenge. For data augmentation, we applied random horizontal and vertical flips during training. Input images are normalized using a mean of `[123.675, 116.28, 103.53]` and a standard deviation of `[58.395, 57.12, 57.375]`.

\paragraph{Massachusetts Roads.} We used the standard split defined in the original paper \cite{mnih2013machine}. Input images were resized to 1532$\times$1532 pixels. The only data augmentation used was random horizontal and vertical flips.

\paragraph{ADE20K.} We used the standard dataset split and followed the classical preprocessing pipeline for this benchmark, which includes resizing, random cropping to 512$\times$512, random horizontal flipping, and normalization.

\subsection{Hyperparameter Configuration}
For full transparency, we provide a detailed summary of the key hyperparameters used for our main PathMamba model on the DeepGlobe dataset.

\paragraph{Backbone (PathMamba).}
The encoder architecture is configured as follows:
\begin{itemize}
    \item \textbf{Architecture:} A four-stage hierarchical backbone with depths of `(2, 2, 8, 2)` and an initial embedding dimension of `96`.
    \item \textbf{Stage Configuration:} Stages 1, 2, and 4 consist of Mamba (VSS) blocks. Stage 3 is our proposed hybrid stage with a sequential `mmmm-aaaa` configuration (4 Mamba blocks followed by 4 Transformer blocks).
    \item \textbf{Regularization:} A drop path rate of `0.2` is applied.
    \item \textbf{Patch Embedding:} A patch size of `4x4` is used.
\end{itemize}

\paragraph{Decoder and Loss Function.}
\begin{itemize}
    \item \textbf{Decoder:} We use the UperNet (`UPerHead`) decoder, which fuses features from all four backbone stages. It employs pyramid pooling with scales of `(1, 2, 3, 6)`.
    \item \textbf{Loss Function:} The total loss is a sum of two components, each with a weight of 1.0: a pixel-wise Focal Loss and a region-based Dice Loss. This hybrid loss is applied to both the main and auxiliary heads.
\end{itemize}

\paragraph{Training and Optimization.}
\begin{itemize}
    \item \textbf{Optimizer:} We use the AdamW optimizer with a learning rate of $6 \times 10^{-5}$, betas of `(0.9, 0.999)`, and a weight decay of `0.01`.
    \item \textbf{Learning Rate Schedule:} A linear warm-up schedule is used for the first 1,500 iterations, increasing the learning rate from $10^{-6}$ to the base learning rate. This is followed by a polynomial decay schedule for the remainder of the training.
    \item \textbf{Training Duration:} All models on the DeepGlobe dataset are trained for a total of 160,000 iterations.
    \item \textbf{Batch Size:} A batch size of 4 per GPU is used.
\end{itemize}

\subsection{Experimental Protocol}
\paragraph{Training Runs and Variation.}
Due to the significant computational cost associated with training large-scale vision models, all experimental results reported in the main paper are from a single, complete training run for each model configuration. This is a common practice in the field for experiments of this scale. As a result, measures of variation such as standard deviation or error bars are not applicable, as they would require multiple training runs which were computationally prohibitive.

\end{document}